\documentclass[pdflatex,sn-mathphys-num]{sn-jnl}
\DeclareGraphicsExtensions{.pdf}

\usepackage{graphicx}%
\usepackage{multirow}%
\usepackage{amsmath,amssymb,amsfonts}%
\usepackage{amsthm}%
\usepackage{mathrsfs}%
\usepackage[title]{appendix}%
\usepackage{xcolor}%
\usepackage{textcomp}%
\usepackage{manyfoot}%
\usepackage{booktabs}%
\usepackage{algorithm}%
\usepackage{algorithmicx}%
\usepackage{algpseudocode}%
\usepackage{listings}%

\theoremstyle{thmstyleone}%
\theoremstyle{thmstyletwo}%

\theoremstyle{thmstylethree}%

\begin{document}

\title[Article Title]{Insights into Artificial Intelligence Virtual Cells}


\author[1]{\fnm{Chengpeng} \sur{Hu}}\email{chengpenghu25@stu.pku.edu.cn}


\author*[1,2,3]{\fnm{Calvin Yu-Chian} \sur{Chen}}\email{cy@pku.edu.cn}

\affil[1]{\orgdiv{School of Chemical Biology and Biotechnology}, \orgname{Peking University Shenzhen Graduate School}, \orgaddress{\street{No. 2199 Lishui Road}, \city{Shenzhen}, \postcode{518055}, \state{Guangdong}, \country{China}}}

\affil*[2]{\orgdiv{School of AI for Science}, \orgname{Peking University}, \orgaddress{\street{No. 5 Yiheyuan Road}, \city{Beijing}, \postcode{100091}, \country{China}}}

\affil[3]{\orgdiv{Department of Medical Research}, \orgname{China Medical University Hospital}, \orgaddress{\street{No. 2, Yude Road, North District}, \city{Taichung}, \postcode{40447}, \country{Taiwan}}}


\abstract{Artificial Intelligence Virtual Cells (AIVCs) aim to learn executable, decision-relevant models of cell state from multimodal, multiscale measurements. Recent studies have introduced single-cell and spatial foundation models, improved cross-modality alignment, scaled perturbation atlases, and explored pathway-level readouts. Nevertheless, although held-out validation is standard practice, evaluations remain predominantly within single datasets and settings; evidence indicates that transport across laboratories and platforms is often limited, that some data splits are vulnerable to leakage and coverage bias, and that dose, time and combination effects are not yet systematically handled. Cross-scale coupling also remains constrained, as anchors linking molecular, cellular and multicellular levels are sparse, and alignment to scientific or clinical readouts varies across studies. We propose a model-agnostic Cell-State Latent (CSL) perspective that organizes learning via an operator grammar: measurement, lift/project for cross-scale coupling, and perturbation for dosing and scheduling. This view motivates a decision-aligned evaluation blueprint across modality, scale, context and perturbation, and emphasizes function-space readouts such as pathway activity, spatial neighborhoods and clinically relevant endpoints. We recommend operator-aware data design, leakage-resistant partitions, and transparent calibration and reporting to enable reproducible, like-for-like comparisons.}




\maketitle  

\section{Introduction}\label{sec1}

Cells exhibit behaviors that arise from interactions across the molecular, organellar, cellular, and multicellular levels. Early efforts to build “virtual cells” demonstrated that it is technically feasible to integrate heterogeneous biological processes into mechanistic simulations of whole cells, such as models of Mycoplasma genitalium and Escherichia coli. However, these rule-based systems relied on hand-specified assumptions, were often brittle to parameter uncertainty and condition shifts, and have been difficult to scale across experimental and biological diversity \citep{karr2012whole,tomita1999cell}. In parallel, recent perspectives emphasize a complementary, data-driven avenue: learning virtual cell models directly from multimodal measurements to infer cellular states and their responses to genetic and chemical perturbations \citep{bunne2024build,qian2025grow}.

The rapid expansion of high-throughput measurement technologies now provides unprecedented opportunities for this approach. Multi-omics datasets are increasingly available across multicellular contexts, but bulk measurements often obscure heterogeneity and dynamics. In contrast, single-cell multi-omics, which spans transcriptomes, genomes, proteomes, and epigenomes, offers a natural substrate for virtual cell modeling, as it captures variation and regulation at the fundamental unit of biological organization \citep{regev2017human, stoeckius2017simultaneous}. Single-cell proteogenomic studies, spatial transcriptomics, and imaging-based assays provide complementary layers of information, making it possible to align intracellular processes with tissue and organ architecture \citep{rodriques2019slide,bray2016cell}. To facilitate large-scale modeling, several reference corpora have been compiled. For example, Tahoe-100M aggregates diverse perturbational single-cell transcriptomes at single-cell resolution, X-Atlas/Orion releases a genome-wide Perturb-seq resource spanning millions of cells with dose-dependent perturbations, and SToCorpus-88M provides tens of millions of spatially resolved cell profiles suitable for pretraining foundation models \citep{zhang2025tahoe,huang2025x,zhao2025stofm}. Together with the Human Cell Atlas and related initiatives \citep{regev2017human}, these corpora enable AI methods that extract coherent representations from multimodal, high-dimensional cellular data.

As large-scale pretraining and multimodal integration have transformed language, vision, and chemistry \citep{brown2020language,jumper2021highly}, similar strategies are now being applied to cellular modeling. Transformer-based genomic language models capture regulatory signals across DNA sequences \citep{consens2025transformers,szalata2024transformers}, while large-scale single-cell models such as scGPT and Geneformer demonstrate that pretraining on millions of cells can produce representations useful for diverse downstream tasks \citep{theodoris2023transfer,cui2024scgpt}. Graph neural networks have been deployed to model cell-cell and gene-gene relationships \citep{kipf2016semi}, and diffusion models are emerging for generative simulation of single-cell and spatial transcriptomics \citep{lotfollahi2020conditional}. These trajectories collectively signal an accelerating convergence of AI and cell biology. At the same time, evaluations reveal uneven performance: model advantages are often task-specific, and in some zero-shot settings simple baselines rival or exceed current foundation models \citep{liu2023evaluating,kedzierska2025zero}. These findings underscore the importance of standardized benchmarks, causal perspectives on dataset bias, and systematic evaluation frameworks to ensure reliability and fairness \citep{jones2024causal}.

Looking ahead, we articulate a model-agnostic organizing framework. Here, we introduce the \textbf{Cell-State Latent (CSL)} as a representational space for encoding heterogeneous measurements and relating modalities and scales. The purpose of CSL is to provide a higher-level organizing view of virtual cells that abstracts away architectural details, enabling cross-modal alignment, linkage across molecular, cellular, and multicellular levels, and perturbational predictions. Viewed through CSL, we suggest simple guidance for unified and effective evaluation. By prioritizing representational principles and shared evaluation standards over any one architecture, CSL aims to anchor the emerging area of Artificial Intelligence Virtual Cells (AIVCs) in reproducible, comparable progress.


\begin{figure*}[t]
  \centering
  \includegraphics[width=\textwidth]{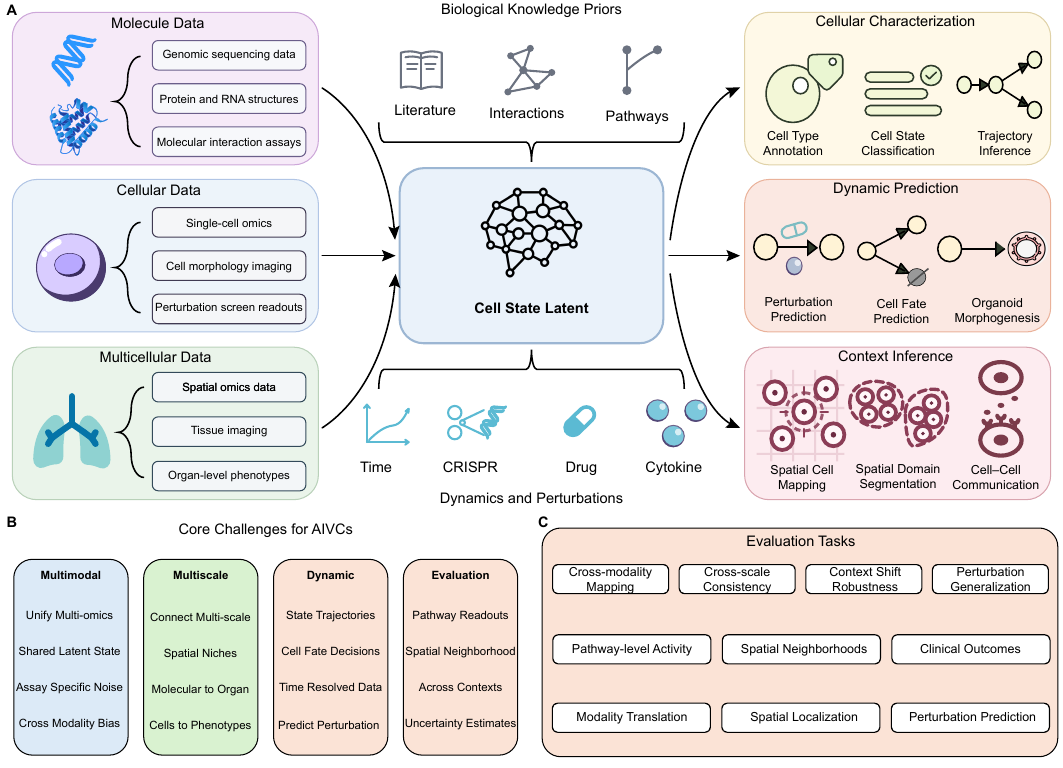}
  \caption{\textbf{Overview of the Artificial Intelligence Virtual Cell (AIVC) framework and the Cell-State Latent (CSL) perspective.}(A) The conceptual architecture of the AIVC. The framework integrates heterogeneous multimodal and multiscale data with Biological Knowledge Priors and Dynamics and Perturbations to learn a unified Cell State Latent (CSL) representation. This shared latent space supports diverse downstream tasks categorized into Cellular Characterization, Dynamic Prediction, and Context Inference.(B) Core challenges for AIVCs.(C) A decision-aligned evaluation blueprint.}
  \label{fig:wide}
\end{figure*}

\section{Core Challenges for AIVCs}\label{sec2}

\subsection{Multimodal Heterogeneity and Unification}\label{sec2.1}

An AIVC must integrate the rapidly expanding single-cell multi-omics landscape to achieve more accurate and comprehensive modeling, making integration both urgent and difficult. Most datasets still measure a single layer per cell or provide only partial pairing, so downstream analysis must align unpaired modalities. Moreover, RNA, protein, and chromatin assays differ in sparsity, dynamic range, and noise, which complicates joint modeling. These constraints have motivated methods that are explicitly missingness-aware and modality-specific in their likelihoods (for example, totalVI, MultiVI) and methods that exploit prior biological knowledge when natural one-to-one pairing is absent (for example, GLUE) \cite{gayoso2021joint,ashuach2023multivi,cao2022multi,luecken2022benchmarking}. We treat each assay as a measurement process acting on an underlying cell state, and we use biologically grounded priors—such as enhancer–gene links, transcription factor regulons, protein–protein interactions, and spatial adjacency—to improve identifiability, sample efficiency, and stability \cite{cao2022multi,fulco2019activity,garcia2019benchmark,szklarczyk2019string,staahl2016visualization}.

\textbf{Single-modality progress.}
For the transcriptome, large-scale pretraining has established strong baselines: scGPT ($\approx$33 million cells) and Geneformer ($\approx$30 million cells) transfer to annotation, perturbation, and network-level tasks, illustrating how much signal expression alone can carry \cite{cui2024scgpt,theodoris2023transfer}. In parallel, language-based formulations such as Cell2Sentence represent expression profiles as “cell sentences” and leverage LLM pretraining for annotation and related downstream tasks, providing a text-aligned alternative measurement operator \cite{levine2024cell2sentence}. For chromatin accessibility, sequence-aware models such as scBasset leverage DNA sequence to denoise and structure scATAC-seq, although interpretation and cross-layer alignment remain sensitive to preprocessing and peak-to-gene linking \cite{yuan2022scbasset,de2024systematic}. For proteins, CITE-seq/REAP-seq established paired mRNA--protein profiling, and multiple studies report gene-specific and often modest RNA--protein concordance at single-cell resolution, motivating either direct inclusion of proteins or explicit cross-modal learning when protein endpoints are central \cite{stoeckius2017simultaneous,reimegaard2021combined}. Collectively, these works quantify the strengths of single-modality encoders while indicating that some targets (for example, protein abundance or regulatory state) are more reliably addressed when multiple modalities are considered jointly.

\textbf{A compact mathematical view.}
Let modality $m$ produce an observation $x_m \in \mathcal{X}_m$. Encoders $f_m:\mathcal{X}_m \to \mathbb{R}^{d}$ map data to a shared state $z\in\mathbb{R}^{d}$. A modality-specific measurement operator $\mathcal{M}_m$ links $z$ to the observation model through a likelihood $p_{\theta_m}\!\big(x_m \mid \mathcal{M}_m(z)\big)$ (for example, over-dispersed counts for RNA; mixture or zero-inflated models for ADTs). This generative component is instantiated in totalVI and MultiVI via amortized variational inference; GLUE augments alignment by incorporating a guidance-graph prior across feature spaces \cite{gayoso2021joint,ashuach2023multivi,cao2022multi,luecken2022benchmarking}. When paired or pseudo-paired samples exist, alignment can also be encouraged by the standard InfoNCE objective (widely used in contrastive learning):

\begin{equation}
\label{eq:infoNCE}
\mathcal{L}_{\text{InfoNCE}}
= - \log \frac{\exp\!\big(s(g_a(z),\,g_b(z^{+}))/\tau\big)}%
{\exp\!\big(s(g_a(z),\,g_b(z^{+}))/\tau\big) + \sum_{k}\exp\!\big(s(g_a(z),\,g_b(z_k^{-}))/\tau\big)} ,
\end{equation}

where $g_a, g_b$ are projection heads and $\tau$ is a temperature (known formula) \cite{oord2018representation,radford2021learning}. For exposition rather than training, we summarize common integrators with the following abstraction:

\begin{equation}
\label{eq:obs}
\mathcal{L}_{\mathrm{obs}}(\theta)
= \sum_{s,m}\,\mathbb{E}\!\left[-\log p_{\theta_{s,m}}\!\left(x_{s,m}\mid \mathcal{M}_{s,m}(z)\right)\right]
\;+\; \lambda_{\mathrm{align}}\,\mathcal{L}_{\mathrm{align}}(\theta)
\;+\; \lambda_{\mathrm{prior}}\,\mathcal{L}_{\mathrm{prior}}(\theta).
\end{equation}

Here $\mathcal{L}_{\mathrm{align}}$ denotes contrastive, distributional, or graph-informed alignment when supervision permits, and $\mathcal{L}_{\mathrm{prior}}$ encodes structured biological knowledge (for example, enhancer–gene links, TF–target regulons, or protein–protein interactions). Setting $\mathcal{L}_{\mathrm{align}}=0$ recovers purely generative VAEs as in totalVI and MultiVI; setting the likelihood term to zero with $\mathcal{L}_{\mathrm{align}}\neq 0$ recovers dual-encoder alignment; choosing $\mathcal{L}_{\mathrm{prior}}\neq 0$ corresponds to prior-regularized integration as in GLUE \cite{gayoso2021joint,ashuach2023multivi,cao2022multi,luecken2022benchmarking,oord2018representation,radford2021learning}.

\textbf{Multimodal integrators.}
Weighted-nearest-neighbors (WNN) in Seurat v4 learns per-cell modality weights and fuses neighborhoods across RNA and ADT, providing a robust, scalable baseline for paired data; its local linear fusion can under-capture nonlinear cross-layer effects \cite{hao2021integrated}. Probabilistic generative models totalVI and MultiVI provide joint latents with modality-specific likelihoods and principled uncertainty, handling missing modalities in practice; sensitivities include likelihood choice and data imbalance at very high dimensions \cite{gayoso2021joint,ashuach2023multivi}. Prior-informed GLUE addresses unpaired integration by bridging feature spaces via a regulatory guidance graph; performance depends on the quality and coverage of priors and graph construction \cite{cao2022multi}. An alternative approach, scJoint, integrates scRNA-seq and scATAC-seq at atlas scale by leveraging transfer learning from large transcriptomic references, and has been shown to improve label transfer across modalities\cite{lin2022scjoint}. Cross-modal translators such as BABEL enable RNA$\leftrightarrow$ATAC (and RNA$\leftrightarrow$protein) prediction to ``virtually pair'' assays, but translation can degrade under domain shift and is not causal by construction \cite{wu2021babel}. Under weak feature linkage, MaxFuse shows that iterative co-embedding and smoothing can improve matching across spatial and sequencing modalities, albeit with additional computational overhead \cite{chen2024integration}. Independent benchmarks across tasks and regimes consistently find that method rankings vary with data scale, pairing pattern, and evaluation metric; no single approach dominates universally, underscoring the value of explicit assumptions and transparent objectives \cite{luecken2022benchmarking,xiao2024benchmarking}.

\textbf{Summary.} We propose an operator-based objective for virtual cells that treats each assay as a likelihood term, couples modalities with an alignment regularizer when paired data exist, and encodes biological structure with interpretable priors, as formalized earlier. Evidence indicates that modality-specific likelihoods and measurement operators preserve assay physics. Cross-modal alignment benefits from contrastive or distributional objectives with paired data, while structured priors help when pairing is limited. Empirical performance remains sensitive to preprocessing and data regime, especially for scATAC-seq where pipeline choices affect integration. Practically, an AIVC should model multiple modalities in a shared latent with reconstruction, feasible alignment, and biologically grounded priors, and assess robustness through ablations and shift stress tests \cite{hao2021integrated,gayoso2021joint,ashuach2023multivi,cao2022multi,luecken2022benchmarking,stoeckius2017simultaneous,chen2024integration,de2024systematic,xiao2024benchmarking,fulco2019activity,garcia2019benchmark,szklarczyk2019string,staahl2016visualization}.

\subsection{Multiscale Structure and Transport}\label{sec2.2}

High-throughput measurements are expanding across molecular, cellular and multicellular levels, with spatially resolved transcriptomics recognized as a Method of the Year and now routine in methods and applications literature \cite{marx2021method,park2022spatial}. At the multicellular scale, large whole-slide imaging corpora and foundation-model pretraining have emerged alongside clinical benchmarks that probe generalization in real cohorts \cite{campanella2025clinical}. Spatial omics resources are being consolidated into cross-organ collections and benchmarking suites that facilitate alignment and integration across platforms and resolutions \cite{park2022spatial,jaume2024hest}. In parallel, molecular-level modeling advances hypothesis generation through structure prediction and genome language models, cellular-level probabilistic modeling improves denoising and uncertainty handling for multi-omics, and multicellular-level studies align dissociated profiles with spatial context and evaluate image-based prediction of gene expression \cite{abramson2024accurate,benegas2025genomic,gayoso2021joint,ashuach2023multivi,wang2025benchmarking}. However, current studies still report variability in cross-cohort robustness for single-scale foundation models \cite{campanella2025clinical,kedzierska2025zero}, sensitivity to distribution shift in medical imaging \cite{castro2020causality}, and resolution or registration mismatches together with platform heterogeneity in spatial assays \cite{park2022spatial,jaume2024hest,wang2025benchmarking}. These considerations motivate joint multiscale modeling that uses cross-level anchors to couple information across resolutions and evaluates predictions with readouts across levels, leveraging emerging multimodal co-registrations that connect morphology, imaging and spatial transcriptomics \cite{park2022spatial,chelebian2025combining,shen2025spatial}.

\textbf{Molecular/cellular progress.} At the molecular level, accurate complex-structure prediction provides physically informed constraints on feasible interactions and perturbation effects \cite{abramson2024accurate}. Genome language models capture sequence regularities and support self-supervised hypothesis generation about regulatory elements \cite{benegas2025genomic}. At the cellular level, transformer-based analyses tailored to single-cell readouts are summarized in recent surveys \cite{szalata2024transformers}. Probabilistic generative frameworks make the observation model explicit: totalVI accounts for RNA/ADT (antibody-derived tags) count noise and background to yield a joint latent, and MultiVI extends to multi-omics with unpaired or missing modalities \cite{gayoso2021joint,ashuach2023multivi}. Complementarily, neighborhood-based fusion learns per-cell modality weights and offers a robust, label-free baseline for multi-omic integration \cite{hao2021integrated}.

\textbf{Multicellular context progress.} Spatial methods provide neighborhood structure and multiresolution context beyond dissociated assays. Tangram maps single-cell references into spatial coordinates \cite{biancalani2021deep}. cell2location and DestVI perform Bayesian deconvolution to estimate cell-type composition and within-type variation from spot-level measurements \cite{kleshchevnikov2022cell2location,lopez2022destvi}. CytoSPACE assigns single cells under limited gene recovery using optimization \cite{vahid2023high}. HE2RNA and SEQUOIA predict gene-expression programs directly from H\&E whole-slide images, linking morphology with molecular readouts \cite{schmauch2020deep,pizurica2024digital}. COVET/ENVI jointly embed spatial and single-cell measurements to represent niche context \cite{haviv2025covariance}. Organ-focused foundation models such as Nephrobase Cell+ integrate sc/snRNA, snATAC, and spatial transcriptomics for cross-scale kidney analyses \cite{li2025nephrobase}. Nevertheless, open issues remain across molecular, cellular, and multicellular levels: at the molecular level, structure and sequence models provide constraints yet do not by themselves resolve causal regulatory effects \textit{in situ} \cite{abramson2024accurate,benegas2025genomic,szalata2024transformers}; at the cellular level, handling missing modalities and transferring across cohorts remains challenging \cite{gayoso2021joint,ashuach2023multivi,kedzierska2025zero,hao2021integrated}; at the multicellular level, resolution and registration mismatches, platform heterogeneity, and distribution shift can hinder robustness and calibration \cite{park2022spatial,jaume2024hest,wang2025benchmarking,castro2020causality}. These considerations motivate the cross-scale coupling and evaluation introduced next.

\textbf{Cross-scale integrators.} Recent work increasingly relates information across biological levels and readouts. One direction learns shared representations across multicellular resolutions using large spatial-omics corpora, exemplified by SToFM trained on SToCorpus-88M \cite{zhao2025stofm}, and another provides specimen-matched anchors by co-registering macroscopic MRI with spatial transcriptomics \cite{shen2025spatial}. Within single cells, translation across modalities supplies supervised mappings that can act as local transport operators; BABEL translates between RNA$\leftrightarrow$ATAC$\leftrightarrow$protein \cite{wu2021babel}. Prior-informed joint embedding uses regulatory graphs to bridge omics spaces, as in GLUE \cite{cao2022multi}. Nicheformer conditions cell representations on spatial neighborhoods with a transformer, coupling single-cell states to niche context and enabling cross-level transfer in spatial omics \cite{TejadaLapuerta2025Nicheformer}. Perturbation-linked models predict morphological responses from transcriptomic perturbations, yielding concrete cross-level readouts for evaluation \cite{wang2025prediction}.

\textbf{Summary.} We summarize cross-scale coupling with an operator-aware objective. Let $\mathcal{M}_{s,m}$ denote scale–modality measurement operators, $\mathcal{L}$ lifting, $\mathcal{P}$ projection, $\Delta$ perturbations, and $z$ the cell-state latent:
\begin{equation}
\label{eq:xscale}
\min_{\theta} J_{\mathrm{xscale}}(\theta)
\;=\; \mathcal{L}_{\mathrm{obs}}(\theta)
\;+\; \lambda_{\mathrm{xscale}}\,\mathcal{L}_{\mathrm{xscale}}(\theta).
\end{equation}
where $\mathcal{L}_{\mathrm{obs}}$ fits modality-appropriate observation models at each scale (for example, count-based likelihoods for single-cell RNA and protein) \cite{gayoso2021joint,ashuach2023multivi}; $\mathcal{L}_{\mathrm{xscale}}$ implements cross-scale checks using available anchors, such as cycle consistency of $\mathcal{P}\!\circ\!\mathcal{L}$ on held-out spatial/slide patches or agreement between $\mathcal{M}_{\mathrm{context}}(\mathcal{P}\,\Delta z)$ and observed multicellular responses \cite{shen2025spatial,zhao2025stofm,wu2021babel,wang2025prediction}; and $\mathcal{L}_{\mathrm{prior}}$ incorporates priors, including structure-informed constraints from AlphaFold~3 and regulatory graphs \cite{abramson2024accurate,cao2022multi}.
 Motivated by these studies, two design considerations are emphasized: operator-aware data collection that introduces sparse but informative cross-level anchors---such as co-assayed spatial patches or small perturbation panels with image readouts---to probe identifiability \cite{chelebian2025combining,shen2025spatial,wang2025prediction}; and evaluation that reports within-scale accuracy together with cross-scale self-consistency and perturbational validity, complemented by shift-aware stress tests where appropriate \cite{wang2025benchmarking,kedzierska2025zero,castro2020causality}.

\subsection{Dynamics and perturbations}\label{sec2.3}

Given a unified latent representation of cell state, this subsection examines temporal evolution and responses to perturbations induced by genetic or chemical agents. Time‐resolved constraints are commonly derived from splicing‐aware RNA velocity and related vector‐field reconstructions, which capture short‐horizon trends yet depend on velocity estimation, sampling density, and population asynchrony \cite{lederer2024statistical,qiu2022mapping}. Direct time supervision further arises from metabolic RNA labeling (4sU/SLAM‐seq and single‐cell variants such as scSLAM‐seq and scNT‐seq), lineage tracing with CRISPR barcodes, and live‐cell reporters with time‐lapse imaging, all of which complement velocity by providing absolute time or ancestry information \cite{herzog2017thiol,erhard2019scslam,qiu2020massively,mckenna2016whole}. Orthogonal phenotypic endpoints from high‐throughput Cell Painting provide morphology‐level readouts for genetic and small‐molecule perturbations and serve as cross‐modality anchors for coherence and robustness analyses \cite{chandrasekaran2024three}.

\textbf{Assays and public resources.} Recent releases broaden coverage and standardization. Tahoe‐100M aggregates on the order of $10^8$ single‐cell profiles across $\sim$1{,}200 small molecules and $\sim$50 lines, supporting large‐scale pretraining and counterfactual testing \cite{zhang2025tahoe}. X‐Atlas/Orion contributes genome‐wide Perturb‐seq ($\sim$8 M cells) via a scalable Fix–Cryopreserve workflow suitable for dose‐ and line‐aware modeling \cite{huang2025x}. Community efforts include the Virtual Cell Challenge, an open, recurring benchmark that provides purpose-built perturbational datasets and a prespecified evaluation pipeline, and the OP3 perturbation-prediction benchmark, which standardizes tasks and metrics for small-molecule response prediction and maintains a public leaderboard \cite{roohani2025virtual,szalata2024benchmark}. For cross‐modality anchoring, JUMP‐CP/CPJUMP1 provides millions of matched images with standardized processing \cite{chandrasekaran2024three}. Data harmonization resources, such as scPerturb and PerturBase, consolidate single‐cell perturbation studies across modalities, lowering barriers for multi‐source training and transport analyses \cite{peidli2024scperturb,wei2025perturbase},.

\textbf{Models and limitations.} Perturbation‐aware virtual‐cell models typically combine: a measurement layer that aligns or reconstructs observations (including RNA$\leftrightarrow$morphology pairs); a dynamics layer that imposes temporal consistency using velocity, labeling, or time‐series constraints; and a perturbation layer that conditions on dose, schedule, and combination. Vector‐field and fate‐mapping frameworks (e.g., dynamo) integrate velocity evidence to reconstruct continuous flows and provide weak supervision for short‐term forecasts; Bayesian and manifold‐constrained variants improve calibration yet remain sensitive to velocity quality and experimental design \cite{lederer2024statistical,qiu2022mapping}. Latent conditional generators, including scGen and CPA, serve as canonical baselines for extrapolating to unseen doses, times, and some combinations from limited supervision \cite{lotfollahi2019scgen,lotfollahi2023predicting}. Causal‐matching methods such as CINEMA‐OT estimates treatment effects under explicit confounding assumptions using optimal transport \cite{dong2023causal}. Work in 2025 emphasizes scale and generalization: STATE introduces an architecture for cross‐context perturbation prediction trained on very large perturbed corpora \cite{adduri2025predicting}; Tahoe-x1 scales perturbation-trained single-cell foundation models on the Tahoe-100M compendium \cite{gandhi2025tahoe,zhang2025tahoe}; large perturbation models (LPM) disentangle perturbation, readout, and context for unified prediction across modalities \cite{miladinovic2025silico}; TxPert couples biochemical and knowledge‐graph structure to improve out‐of‐distribution prediction for single and double genetic perturbations across cell lines \cite{wenkel2025txpert}. On morphological endpoints, IMPA employs generative style‐transfer to predict imaging‐level responses to genetic and chemical perturbations, and CellFlow applies flow matching to model distribution‐to‐distribution morphology shifts conditioned on perturbations \cite{palma2025predicting,zhang2025cellflux}. Complementary methodological studies target measurement and evaluation: PS quantifies heterogeneous single‐cell responses within perturbed populations to separate effect size from dispersion, and TRADE provides transcriptome‐wide statistics tailored to large perturbation atlases, facilitating principled differential analyses at scale \cite{song2025decoding,nadig2025transcriptome}. Across model families, recurring limitations include fragile composition for unseen combinations or schedules without explicit biological structure, limited testing of order sensitivity between aggregation/measurement and perturbation even when paired readouts exist, and incomplete transport across donors, laboratories, and instruments despite multi‐site datasets and public leaderboards \cite{chandrasekaran2024three,roohani2025virtual,szalata2024benchmark,lotfollahi2023predicting,wenkel2025txpert}.

\textbf{Summary.} We propose a compact objective that separates observation, temporal, and perturbation signals, consistent with the notation used elsewhere:
\begin{equation}
\label{eq:pert}
\min_{\theta} J_{\mathrm{pert}}(\theta)
\;=\; \mathcal{L}_{\mathrm{obs}}(\theta)
\;+\; \lambda_{\mathrm{xscale}}\,\mathcal{L}_{\mathrm{xscale}}(\theta)
\;+\; \lambda_{\mathrm{pert}}\,\mathcal{L}_{\mathrm{pert}}(\theta).
\end{equation}
Here, $\mathcal{L}_{\mathrm{obs}}$ fits cross-modality and cross-scale observations; $\mathcal{L}_{\mathrm{xscale}}$ enforces temporal and cross-level coherence using velocity, metabolic or lineage labeling, and time-series supervision; and $\mathcal{L}_{\mathrm{pert}}$ fits responses to perturbations specified by dose, duration, and combinations. In this view, velocity/labeling/time-series data supervise $\mathcal{L}_{\mathrm{xscale}}$ \cite{lederer2024statistical,qiu2022mapping,herzog2017thiol,erhard2019scslam,qiu2020massively,mckenna2016whole}; latent conditional generators such as scGen/CPA and causal matching as in CINEMA-OT supervise $\mathcal{L}_{\mathrm{pert}}$
 \cite{lotfollahi2019scgen,lotfollahi2023predicting,dong2023causal}; and large‐scale morphological perturbation imaging resources (e.g., JUMP‐CP/CPJUMP1) are not one‐to‐one image–transcriptome pairs but provide phenotype‐level distributions and perturbation readouts that complement training and evaluation \cite{chandrasekaran2024three}. Large perturbation atlases and standardized benchmarks provide regimes for training and transport assessment \cite{zhang2025tahoe,huang2025x,roohani2025virtual,szalata2024benchmark}.

\subsection{Evaluation}\label{sec2.4}

Evaluation for an AIVC asks whether a model supports dependable scientific and translational decisions across laboratories, protocols, and populations. In the operator view introduced earlier (measurement at each scale, lift and project between scales, and explicit perturbation operators) the readout should extend beyond gene- or cell-level errors to pathway activity, spatial neighborhood structure, and clinically relevant endpoints. The aim is to connect modeling choices to measurable biological or clinical benefit, with stress tests under distribution shift, leakage-resistant data partitions, and, where available, external cohorts \cite{bunne2024build,qian2025grow,campanella2025clinical,wang2025benchmarking,yang2024limits,ktena2024generative,kretschmer2025coverage,joeres2025data}.

\textbf{Evaluation axes and representative tasks.} We organize evaluation along four axes implied by the operator grammar. These axes instantiate the operators introduced earlier ($\mathcal{M}$, $\mathcal{L}/\mathcal{P}$, context $c$, and perturbation $\Delta$) so that evaluation mirrors model design. (i) Cross-modality mapping evaluates alignment between single-cell profiles (for example, RNA/ATAC) and spatial or histology data through cell-type localization accuracy, gene-level reconstruction/correlation, and neighborhood/niche consistency. Recent studies add platform-aware comparisons and external validation; image-to-transcriptome and pathology foundation models extend these readouts to clinical endpoints at institutions not seen during development \cite{pizurica2024digital,chen2024towards,xu2024whole,campanella2025clinical,wang2025benchmarking,you2024systematic,haviv2025covariance}. (ii) Cross-scale consistency asks whether lift/project operations preserve structure across molecular, cellular, and multicellular levels. Multicellular-scale corpora and spatial omics enable tests of whether dissociated single-cell profiles can be localized and whether image-based predictors recover gene or pathway signals across slides and cohorts; while widely used proxies exist, rigorous cross-scale scoring remains an active area of method development \cite{pizurica2024digital,chen2024towards,xu2024whole,campanella2025clinical,wang2025benchmarking,you2024systematic}. (iii) Context shift probes transport across sites, acquisition protocols, and populations. Clinical-facing pathology benchmarks and spatial prediction studies increasingly report external-center performance and analyze robustness to cohort heterogeneity \cite{xu2024whole,campanella2025clinical,wang2025benchmarking,yang2024limits,ktena2024generative}. (iv) Perturbation generalization focuses on perturbations and dynamics. Harmonized resources and standardized tasks quantify generalization across cell backgrounds, chemical scaffolds, doses, and combinations using gene-wise error and pathway concordance; morphology-based screening adds distributional image metrics and prospective gains in hit-rate and diversity in selected settings. For dynamics and lineage, vector-field constraints and time-labeling protocols are compared, with reporting of uncertainty and biological plausibility emphasized alongside fit \cite{chandrasekaran2024three,moshkov2024learning,fredin2024cell,peidli2024scperturb,wei2025perturbase,szalata2024benchmark,lederer2024statistical,zhang2025benchmarking}.

\textbf{Frequent failure modes.} First, metric–goal mismatch persists: strong within-slide correlations may not translate into prognostic or diagnostic benefit, motivating studies that tie spatial or image-based prediction to survival, biomarker, or clinical endpoints \cite{campanella2025clinical,wang2025benchmarking}. Second, distribution shift and information leakage inflate optimism: center-specific acquisition, scaffold or donor overlap, and near partitions can hide brittleness; recent work quantifies coverage bias in small-molecule learning and proposes principled splitters that reduce similarity across partitions \cite{kretschmer2025coverage,joeres2025data}. Third, causal identifiability is often under-served: association under a fixed protocol can confound perturbation effects; emerging causal and dynamics benchmarks begin to address this gap, and fairness-oriented studies highlight additional risks at deployment \cite{yang2024limits,ktena2024generative,chevalley2025large}.

\textbf{Unified scoring.} To align evaluation with training while keeping them decoupled, we score models in a task-relevant function space. Let $\phi$ map predictions and observations to decision-relevant readouts (for example, pathway activity, neighborhood distances, calibration or clinical endpoints). For each stress-test distribution $P_k$ along an axis $k \in \{\text{modality}, \text{scale}, \text{context}, \text{perturbation}\}$, define a task loss $\mathcal{L}^{(k)}(\cdot,\cdot)$ on readouts and aggregate
\begin{equation}
\label{eq:risk}
\mathcal{R}(\theta)
\;=\;\sum_{k}\alpha_k\,\mathbb{E}_{(x,c,a,y)\sim P_k}\!\left[
\mathcal{L}^{(k)}\!\big(\phi(f_\theta(x,c,a)),\,\phi(y)\big)
\right].
\end{equation}
Here $c$ denotes site/protocol/population context and $a$ denotes perturbations (targets, doses, combinations); the weights satisfy $\alpha_k\!\ge\!0$ and $\sum_k\alpha_k=1$. This formulation specifies an aggregate risk for reporting that is consistent with the operator notation used elsewhere, without prescribing a particular training objective.

\textbf{Summary.} Building on the aggregate risk defined above and the task-relevant readout function, two directions emerge. First, construct decision-aligned readouts by combining pathway activity scores, neighborhood distances, and uncertainty/calibration diagnostics, then aggregate results across stress-test distributions with weights reflecting end-use priorities; evidence from external-cohort pathology and spatial prediction suggests that such function-space readouts can reorder model rankings and better reflect clinical transportability \cite{pizurica2024digital,chen2024towards,xu2024whole,campanella2025clinical,wang2025benchmarking,yang2024limits,ktena2024generative}. Second, make the stress tests explicit by parameterizing distributional shift through partition similarity and coverage (donor/site for single-cell and spatial data, scaffold for chemistry, and time/dose/combination for perturbations) and examine how the aggregate risk varies with these parameters; recent analyses of coverage bias and leakage-aware splitting offer practical proxies \cite{kretschmer2025coverage,joeres2025data}. Public perturbation and morphology resources (CPJUMP1, scPerturb, PerturBase, OP3) already instantiate the perturbation axis and enable estimation of cross-background and cross-scaffold components of performance \cite{chandrasekaran2024three,moshkov2024learning,fredin2024cell,peidli2024scperturb,wei2025perturbase,szalata2024benchmark}.

\section{Conclusion and outlook}\label{sec3}

Recent advances in multimodal single-cell and spatial profiling have begun to connect molecular signals with cellular neighborhoods and multicellular context. Modeling increasingly considers temporal labeling and perturbation responses. Evaluation is moving beyond reconstruction toward decision-relevant readouts. Despite this progress, four obstacles remain salient: (i) Unification often stops at embedding alignment without assay-aware measurement models or explicit biological priors, so cross-modality coherence is not auditable and can drift from assay physics; (ii) In the absence of within-specimen anchors that link molecular, cellular, and multicellular levels, lift and project operators are weakly identified and cross-level self-consistency is rarely reported; (iii) Time and action are not jointly observed in most settings, leaving intervention operators and composition rules for dose, schedule, and combination under-specified and counterfactual predictions weakly validated; (iv) Scoring still leans on proxy metrics and single-split summaries rather than function-space readouts and shift-parameterized risk, and calibration and uncertainty are not treated as first-class outputs \cite{jones2024causal, chen2025giotto,bunne2024build,qian2025grow}.

In this context, we use CSL as a shared language rather than a recipe. It specifies a cell state space and operators for measurement, lifting, projection, and perturbation, attaching assays to a common representation that links modalities, scales, and interventions. This separates representation from architecture and aligns objectives with evaluation, making claims auditable and conclusions transport across modalities, scales, contexts, and perturbations.

Looking ahead, virtual cells are poised to unify multimodal and multiscale representations, tracing the effects of genetic and pharmacologic perturbations from molecular changes to tissue, organ, and human phenotypes. As perturbation corpora and spatial atlases expand, these models can progress from virtual experiments to counterfactual phenotyping that forecasts efficacy and toxicity with calibrated uncertainty. This trajectory points toward applications in drug discovery and clinical care, from target selection and indication expansion to dosing, combination design, and patient digital twins.


\bibliography{sn-bibliography}

@article{karr2012whole,
  title={A whole-cell computational model predicts phenotype from genotype},
  author={Karr, Jonathan R and Sanghvi, Jayodita C and Macklin, Derek N and Gutschow, Miriam V and Jacobs, Jared M and Bolival, Benjamin and Assad-Garcia, Nacyra and Glass, John I and Covert, Markus W},
  journal={Cell},
  volume={150},
  number={2},
  pages={389--401},
  year={2012},
  publisher={Elsevier}
}

@article{tomita1999cell,
  title={E-CELL: software environment for whole-cell simulation.},
  author={Tomita, Masaru and Hashimoto, Kenta and Takahashi, Koichi and Shimizu, Thomas Simon and Matsuzaki, Yuri and Miyoshi, Fumihiko and Saito, Kanako and Tanida, Sakura and Yugi, Katsuyuki and Venter, J Craig and others},
  journal={Bioinformatics (Oxford, England)},
  volume={15},
  number={1},
  pages={72--84},
  year={1999}
}

@article{bunne2024build,
  title={How to build the virtual cell with artificial intelligence: Priorities and opportunities},
  author={Bunne, Charlotte and Roohani, Yusuf and Rosen, Yanay and Gupta, Ankit and Zhang, Xikun and Roed, Marcel and Alexandrov, Theo and AlQuraishi, Mohammed and Brennan, Patricia and Burkhardt, Daniel B and others},
  journal={Cell},
  volume={187},
  number={25},
  pages={7045--7063},
  year={2024},
  publisher={Elsevier}
}

@article{qian2025grow,
  title={Grow AI virtual cells: three data pillars and closed-loop learning},
  author={Qian, Liujia and Dong, Zhen and Guo, Tiannan},
  journal={Cell Research},
  pages={1--3},
  year={2025},
  publisher={Springer Nature Singapore Singapore}
}

@article{regev2017human,
  title={The human cell atlas},
  author={Regev, Aviv and Teichmann, Sarah A and Lander, Eric S and Amit, Ido and Benoist, Christophe and Birney, Ewan and Bodenmiller, Bernd and Campbell, Peter and Carninci, Piero and Clatworthy, Menna and others},
  journal={elife},
  volume={6},
  pages={e27041},
  year={2017},
  publisher={eLife Sciences Publications, Ltd}
}

@article{stoeckius2017simultaneous,
  title={Simultaneous epitope and transcriptome measurement in single cells},
  author={Stoeckius, Marlon and Hafemeister, Christoph and Stephenson, William and Houck-Loomis, Brian and Chattopadhyay, Pratip K and Swerdlow, Harold and Satija, Rahul and Smibert, Peter},
  journal={Nature methods},
  volume={14},
  number={9},
  pages={865--868},
  year={2017},
  publisher={Nature Publishing Group US New York}
}

@article{rodriques2019slide,
  title={Slide-seq: A scalable technology for measuring genome-wide expression at high spatial resolution},
  author={Rodriques, Samuel G and Stickels, Robert R and Goeva, Aleksandrina and Martin, Carly A and Murray, Evan and Vanderburg, Charles R and Welch, Joshua and Chen, Linlin M and Chen, Fei and Macosko, Evan Z},
  journal={Science},
  volume={363},
  number={6434},
  pages={1463--1467},
  year={2019},
  publisher={American Association for the Advancement of Science}
}

@article{bray2016cell,
  title={Cell Painting, a high-content image-based assay for morphological profiling using multiplexed fluorescent dyes},
  author={Bray, Mark-Anthony and Singh, Shantanu and Han, Han and Davis, Chadwick T and Borgeson, Blake and Hartland, Cathy and Kost-Alimova, Maria and Gustafsdottir, Sigrun M and Gibson, Christopher C and Carpenter, Anne E},
  journal={Nature protocols},
  volume={11},
  number={9},
  pages={1757--1774},
  year={2016},
  publisher={Nature Publishing Group UK London}
}

@article{zhang2025tahoe,
  title={Tahoe-100m: A giga-scale single-cell perturbation atlas for context-dependent gene function and cellular modeling},
  author={Zhang, Jesse and Ubas, Airol A and de Borja, Richard and Svensson, Valentine and Thomas, Nicole and Thakar, Neha and Lai, Ian and Winters, Aidan and Khan, Umair and Jones, Matthew G and others},
  journal={BioRxiv},
  pages={2025--02},
  year={2025},
  publisher={Cold Spring Harbor Laboratory}
}

@article{huang2025x,
  title={X-Atlas/Orion: Genome-wide Perturb-seq Datasets via a Scalable Fix-Cryopreserve Platform for Training Dose-Dependent Biological Foundation Models},
  author={Huang, Ann C and Hsieh, Tsung-Han S and Zhu, Jiang and Michuda, Jackson and Teng, Ashton and Kim, Soohong and Rumsey, Elizabeth M and Lam, Sharon K and Anigbogu, Ikenna and Wright, Philip and others},
  journal={bioRxiv},
  pages={2025--06},
  year={2025},
  publisher={Cold Spring Harbor Laboratory}
}

@article{zhao2025stofm,
  title={SToFM: a Multi-scale Foundation Model for Spatial Transcriptomics},
  author={Zhao, Suyuan and Luo, Yizhen and Yang, Ganbo and Zhong, Yan and Zhou, Hao and Nie, Zaiqing},
  journal={arXiv preprint arXiv:2507.11588},
  year={2025}
}

@article{brown2020language,
  title={Language models are few-shot learners},
  author={Brown, Tom and Mann, Benjamin and Ryder, Nick and Subbiah, Melanie and Kaplan, Jared D and Dhariwal, Prafulla and Neelakantan, Arvind and Shyam, Pranav and Sastry, Girish and Askell, Amanda and others},
  journal={Advances in neural information processing systems},
  volume={33},
  pages={1877--1901},
  year={2020}
}

@article{jumper2021highly,
  title={Highly accurate protein structure prediction with AlphaFold},
  author={Jumper, John and Evans, Richard and Pritzel, Alexander and Green, Tim and Figurnov, Michael and Ronneberger, Olaf and Tunyasuvunakool, Kathryn and Bates, Russ and {\v{Z}}{\'\i}dek, Augustin and Potapenko, Anna and others},
  journal={nature},
  volume={596},
  number={7873},
  pages={583--589},
  year={2021},
  publisher={Nature Publishing Group UK London}
}

@article{consens2025transformers,
  title={Transformers and genome language models},
  author={Consens, Micaela E and Dufault, Cameron and Wainberg, Michael and Forster, Duncan and Karimzadeh, Mehran and Goodarzi, Hani and Theis, Fabian J and Moses, Alan and Wang, Bo},
  journal={Nature Machine Intelligence},
  pages={1--17},
  year={2025},
  publisher={Nature Publishing Group UK London}
}

@article{szalata2024transformers,
  title={Transformers in single-cell omics: a review and new perspectives},
  author={Sza{\l}ata, Artur and Hrovatin, Karin and Becker, S{\"o}ren and Tejada-Lapuerta, Alejandro and Cui, Haotian and Wang, Bo and Theis, Fabian J},
  journal={Nature methods},
  volume={21},
  number={8},
  pages={1430--1443},
  year={2024},
  publisher={Nature Publishing Group US New York}
}

@article{theodoris2023transfer,
  title={Transfer learning enables predictions in network biology},
  author={Theodoris, Christina V and Xiao, Ling and Chopra, Anant and Chaffin, Mark D and Al Sayed, Zeina R and Hill, Matthew C and Mantineo, Helene and Brydon, Elizabeth M and Zeng, Zexian and Liu, X Shirley and others},
  journal={Nature},
  volume={618},
  number={7965},
  pages={616--624},
  year={2023},
  publisher={Nature Publishing Group UK London}
}

@article{cui2024scgpt,
  title={scGPT: toward building a foundation model for single-cell multi-omics using generative AI},
  author={Cui, Haotian and Wang, Chloe and Maan, Hassaan and Pang, Kuan and Luo, Fengning and Duan, Nan and Wang, Bo},
  journal={Nature methods},
  volume={21},
  number={8},
  pages={1470--1480},
  year={2024},
  publisher={Nature Publishing Group US New York}
}

@article{kipf2016semi,
  title={Semi-supervised classification with graph convolutional networks},
  author={Kipf, TN},
  journal={arXiv preprint arXiv:1609.02907},
  year={2016}
}

@article{lotfollahi2020conditional,
  title={Conditional out-of-distribution generation for unpaired data using transfer VAE},
  author={Lotfollahi, Mohammad and Naghipourfar, Mohsen and Theis, Fabian J and Wolf, F Alexander},
  journal={Bioinformatics},
  volume={36},
  number={Supplement\_2},
  pages={i610--i617},
  year={2020},
  publisher={Oxford University Press}
}

@article{liu2023evaluating,
  title={Evaluating the utilities of foundation models in single-cell data analysis},
  author={Liu, Tianyu and Li, Kexing and Wang, Yuge and Li, Hongyu and Zhao, Hongyu},
  journal={bioRxiv},
  pages={2023--09},
  year={2023},
  publisher={Cold Spring Harbor Laboratory}
}

@article{kedzierska2025zero,
  title={Zero-shot evaluation reveals limitations of single-cell foundation models},
  author={Kedzierska, Kasia Z and Crawford, Lorin and Amini, Ava P and Lu, Alex X},
  journal={Genome Biology},
  volume={26},
  number={1},
  pages={101},
  year={2025},
  publisher={Springer}
}

@article{jones2024causal,
  title={A causal perspective on dataset bias in machine learning for medical imaging},
  author={Jones, Charles and Castro, Daniel C and De Sousa Ribeiro, Fabio and Oktay, Ozan and McCradden, Melissa and Glocker, Ben},
  journal={Nature Machine Intelligence},
  volume={6},
  number={2},
  pages={138--146},
  year={2024},
  publisher={Nature Publishing Group UK London}
}

@article{hao2021integrated,
  title={Integrated analysis of multimodal single-cell data},
  author={Hao, Yuhan and Hao, Stephanie and Andersen-Nissen, Erica and Mauck, William M and Zheng, Shiwei and Butler, Andrew and Lee, Maddie J and Wilk, Aaron J and Darby, Charlotte and Zager, Michael and others},
  journal={Cell},
  volume={184},
  number={13},
  pages={3573--3587},
  year={2021},
  publisher={Elsevier}
}

@article{gayoso2021joint,
  title={Joint probabilistic modeling of single-cell multi-omic data with totalVI},
  author={Gayoso, Adam and Steier, Zo{\"e} and Lopez, Romain and Regier, Jeffrey and Nazor, Kristopher L and Streets, Aaron and Yosef, Nir},
  journal={Nature methods},
  volume={18},
  number={3},
  pages={272--282},
  year={2021},
  publisher={Nature Publishing Group US New York}
}

@article{ashuach2023multivi,
  title={MultiVI: deep generative model for the integration of multimodal data},
  author={Ashuach, Tal and Gabitto, Mariano I and Koodli, Rohan V and Saldi, Giuseppe-Antonio and Jordan, Michael I and Yosef, Nir},
  journal={Nature Methods},
  volume={20},
  number={8},
  pages={1222--1231},
  year={2023},
  publisher={Nature Publishing Group US New York}
}

@article{cao2022multi,
  title={Multi-omics single-cell data integration and regulatory inference with graph-linked embedding},
  author={Cao, Zhi-Jie and Gao, Ge},
  journal={Nature Biotechnology},
  volume={40},
  number={10},
  pages={1458--1466},
  year={2022},
  publisher={Nature Publishing Group US New York}
}

@article{luecken2022benchmarking,
  title={Benchmarking atlas-level data integration in single-cell genomics},
  author={Luecken, Malte D and B{\"u}ttner, Maren and Chaichoompu, Kridsadakorn and Danese, Anna and Interlandi, Marta and M{\"u}ller, Michaela F and Strobl, Daniel C and Zappia, Luke and Dugas, Martin and Colom{\'e}-Tatch{\'e}, Maria and others},
  journal={Nature methods},
  volume={19},
  number={1},
  pages={41--50},
  year={2022},
  publisher={Nature Publishing Group US New York}
}

@article{yuan2022scbasset,
  title={scBasset: sequence-based modeling of single-cell ATAC-seq using convolutional neural networks},
  author={Yuan, Han and Kelley, David R},
  journal={Nature Methods},
  volume={19},
  number={9},
  pages={1088--1096},
  year={2022},
  publisher={Nature Publishing Group US New York}
}

@article{reimegaard2021combined,
  title={A combined approach for single-cell mRNA and intracellular protein expression analysis},
  author={Reimeg{\aa}rd, Johan and Tarbier, Marcel and Danielsson, Marcus and Schuster, Jens and Baskaran, Sathishkumar and Panagiotou, Styliani and Dahl, Niklas and Friedl{\"a}nder, Marc R and Gallant, Caroline J},
  journal={Communications biology},
  volume={4},
  number={1},
  pages={624},
  year={2021},
  publisher={Nature Publishing Group UK London}
}

@article{oord2018representation,
  title={Representation learning with contrastive predictive coding},
  author={Oord, Aaron van den and Li, Yazhe and Vinyals, Oriol},
  journal={arXiv preprint arXiv:1807.03748},
  year={2018}
}

@inproceedings{radford2021learning,
  title={Learning transferable visual models from natural language supervision},
  author={Radford, Alec and Kim, Jong Wook and Hallacy, Chris and Ramesh, Aditya and Goh, Gabriel and Agarwal, Sandhini and Sastry, Girish and Askell, Amanda and Mishkin, Pamela and Clark, Jack and others},
  booktitle={International conference on machine learning},
  pages={8748--8763},
  year={2021},
  organization={PmLR}
}

@article{wu2021babel,
  title={BABEL enables cross-modality translation between multiomic profiles at single-cell resolution},
  author={Wu, Kevin E and Yost, Kathryn E and Chang, Howard Y and Zou, James},
  journal={Proceedings of the National Academy of Sciences},
  volume={118},
  number={15},
  pages={e2023070118},
  year={2021},
  publisher={National Academy of Sciences}
}

@article{chen2024integration,
  title={Integration of spatial and single-cell data across modalities with weakly linked features},
  author={Chen, Shuxiao and Zhu, Bokai and Huang, Sijia and Hickey, John W and Lin, Kevin Z and Snyder, Michael and Greenleaf, William J and Nolan, Garry P and Zhang, Nancy R and Ma, Zongming},
  journal={Nature Biotechnology},
  volume={42},
  number={7},
  pages={1096--1106},
  year={2024},
  publisher={Nature Publishing Group US New York}
}

@article{de2024systematic,
  title={Systematic benchmarking of single-cell ATAC-sequencing protocols},
  author={De Rop, Florian V and Hulselmans, Gert and Flerin, Chris and Soler-Vila, Paula and Rafels, Albert and Christiaens, Valerie and Gonzalez-Blas, Carmen Bravo and Marchese, Domenica and Caratu, Ginevra and Poovathingal, Suresh and others},
  journal={Nature biotechnology},
  volume={42},
  number={6},
  pages={916--926},
  year={2024},
  publisher={Nature Publishing Group US New York}
}

@article{xiao2024benchmarking,
  title={Benchmarking multi-omics integration algorithms across single-cell RNA and ATAC data},
  author={Xiao, Chuxi and Chen, Yixin and Meng, Qiuchen and Wei, Lei and Zhang, Xuegong},
  journal={Briefings in Bioinformatics},
  volume={25},
  number={2},
  year={2024},
  publisher={Oxford Academic}
}

@article{lin2022scjoint,
  title={scJoint integrates atlas-scale single-cell RNA-seq and ATAC-seq data with transfer learning},
  author={Lin, Yingxin and Wu, Tung-Yu and Wan, Sheng and Yang, Jean YH and Wong, Wing H and Wang, YX Rachel},
  journal={Nature biotechnology},
  volume={40},
  number={5},
  pages={703--710},
  year={2022},
  publisher={Nature Publishing Group US New York}
}

@article{fulco2019activity,
  title={Activity-by-contact model of enhancer--promoter regulation from thousands of CRISPR perturbations},
  author={Fulco, Charles P and Nasser, Joseph and Jones, Thouis R and Munson, Glen and Bergman, Drew T and Subramanian, Vidya and Grossman, Sharon R and Anyoha, Rockwell and Doughty, Benjamin R and Patwardhan, Tejal A and others},
  journal={Nature genetics},
  volume={51},
  number={12},
  pages={1664--1669},
  year={2019},
  publisher={Nature Publishing Group US New York}
}

@article{garcia2019benchmark,
  title={Benchmark and integration of resources for the estimation of human transcription factor activities},
  author={Garcia-Alonso, Luz and Holland, Christian H and Ibrahim, Mahmoud M and Turei, Denes and Saez-Rodriguez, Julio},
  journal={Genome research},
  volume={29},
  number={8},
  pages={1363--1375},
  year={2019},
  publisher={Cold Spring Harbor Lab}
}

@article{szklarczyk2019string,
  title={STRING v11: protein--protein association networks with increased coverage, supporting functional discovery in genome-wide experimental datasets},
  author={Szklarczyk, Damian and Gable, Annika L and Lyon, David and Junge, Alexander and Wyder, Stefan and Huerta-Cepas, Jaime and Simonovic, Milan and Doncheva, Nadezhda T and Morris, John H and Bork, Peer and others},
  journal={Nucleic acids research},
  volume={47},
  number={D1},
  pages={D607--D613},
  year={2019},
  publisher={Oxford University Press}
}

@article{staahl2016visualization,
  title={Visualization and analysis of gene expression in tissue sections by spatial transcriptomics},
  author={St{\aa}hl, Patrik L and Salm{\'e}n, Fredrik and Vickovic, Sanja and Lundmark, Anna and Navarro, Jos{\'e} Fern{\'a}ndez and Magnusson, Jens and Giacomello, Stefania and Asp, Michaela and Westholm, Jakub O and Huss, Mikael and others},
  journal={Science},
  volume={353},
  number={6294},
  pages={78--82},
  year={2016},
  publisher={American Association for the Advancement of Science}
}

@article{marx2021method,
  title={Method of the Year: spatially resolved transcriptomics},
  author={Marx, Vivien},
  journal={Nature methods},
  volume={18},
  number={1},
  pages={9--14},
  year={2021},
  publisher={Nature Publishing Group US New York}
}

@article{park2022spatial,
  title={Spatial omics technologies at multimodal and single cell/subcellular level},
  author={Park, Jiwoon and Kim, Junbum and Lewy, Tyler and Rice, Charles M and Elemento, Olivier and Rendeiro, Andr{\'e} F and Mason, Christopher E},
  journal={Genome biology},
  volume={23},
  number={1},
  pages={256},
  year={2022},
  publisher={Springer}
}

@article{campanella2025clinical,
  title={A clinical benchmark of public self-supervised pathology foundation models},
  author={Campanella, Gabriele and Chen, Shengjia and Singh, Manbir and Verma, Ruchika and Muehlstedt, Silke and Zeng, Jennifer and Stock, Aryeh and Croken, Matt and Veremis, Brandon and Elmas, Abdulkadir and others},
  journal={Nature Communications},
  volume={16},
  number={1},
  pages={3640},
  year={2025},
  publisher={Nature Publishing Group UK London}
}

@article{jaume2024hest,
  title={Hest-1k: A dataset for spatial transcriptomics and histology image analysis},
  author={Jaume, Guillaume and Doucet, Paul and Song, Andrew and Lu, Ming Yang and Almagro P{\'e}rez, Cristina and Wagner, Sophia and Vaidya, Anurag and Chen, Richard and Williamson, Drew and Kim, Ahrong and others},
  journal={Advances in Neural Information Processing Systems},
  volume={37},
  pages={53798--53833},
  year={2024}
}

@article{abramson2024accurate,
  title={Accurate structure prediction of biomolecular interactions with AlphaFold 3},
  author={Abramson, Josh and Adler, Jonas and Dunger, Jack and Evans, Richard and Green, Tim and Pritzel, Alexander and Ronneberger, Olaf and Willmore, Lindsay and Ballard, Andrew J and Bambrick, Joshua and others},
  journal={Nature},
  volume={630},
  number={8016},
  pages={493--500},
  year={2024},
  publisher={Nature Publishing Group UK London}
}

@article{benegas2025genomic,
  title={Genomic language models: opportunities and challenges},
  author={Benegas, Gonzalo and Ye, Chengzhong and Albors, Carlos and Li, Jianan Canal and Song, Yun S},
  journal={Trends in Genetics},
  year={2025},
  publisher={Elsevier}
}

@article{wang2025benchmarking,
  title={Benchmarking the translational potential of spatial gene expression prediction from histology},
  author={Wang, Chuhan and Chan, Adam S and Fu, Xiaohang and Ghazanfar, Shila and Kim, Jinman and Patrick, Ellis and Yang, Jean YH},
  journal={Nature Communications},
  volume={16},
  number={1},
  pages={1544},
  year={2025},
  publisher={Nature Publishing Group UK London}
}

@article{castro2020causality,
  title={Causality matters in medical imaging},
  author={Castro, Daniel C and Walker, Ian and Glocker, Ben},
  journal={Nature Communications},
  volume={11},
  number={1},
  pages={3673},
  year={2020},
  publisher={Nature Publishing Group UK London}
}

@article{chelebian2025combining,
  title={Combining spatial transcriptomics with tissue morphology},
  author={Chelebian, Eduard and Avenel, Christophe and W{\"a}hlby, Carolina},
  journal={Nature Communications},
  volume={16},
  number={1},
  pages={4452},
  year={2025},
  publisher={Nature Publishing Group UK London}
}

@article{shen2025spatial,
  title={A spatial imaging-transcriptomics paradigm for deciphering the molecular basis of microscopic MRI in the normal brain and Alzheimer’s disease},
  author={Shen, Yiqi and Shen, Yao and Wang, Menglei and Jin, Kaiyu and Yang, Penghui and Cao, Zuozhen and Zhu, Qinfeng and Zhao, Zhiyong and Li, Haotian and Han, Lei and others},
  journal={Cell Reports},
  volume={44},
  number={8},
  year={2025},
  publisher={Elsevier}
}

@article{biancalani2021deep,
  title={Deep learning and alignment of spatially resolved single-cell transcriptomes with Tangram},
  author={Biancalani, Tommaso and Scalia, Gabriele and Buffoni, Lorenzo and Avasthi, Raghav and Lu, Ziqing and Sanger, Aman and Tokcan, Neriman and Vanderburg, Charles R and Segerstolpe, {\AA}sa and Zhang, Meng and others},
  journal={Nature methods},
  volume={18},
  number={11},
  pages={1352--1362},
  year={2021},
  publisher={Nature Publishing Group US New York}
}

@article{kleshchevnikov2022cell2location,
  title={Cell2location maps fine-grained cell types in spatial transcriptomics},
  author={Kleshchevnikov, Vitalii and Shmatko, Artem and Dann, Emma and Aivazidis, Alexander and King, Hamish W and Li, Tong and Elmentaite, Rasa and Lomakin, Artem and Kedlian, Veronika and Gayoso, Adam and others},
  journal={Nature biotechnology},
  volume={40},
  number={5},
  pages={661--671},
  year={2022},
  publisher={Nature Publishing Group US New York}
}

@article{lopez2022destvi,
  title={DestVI identifies continuums of cell types in spatial transcriptomics data},
  author={Lopez, Romain and Li, Baoguo and Keren-Shaul, Hadas and Boyeau, Pierre and Kedmi, Merav and Pilzer, David and Jelinski, Adam and Yofe, Ido and David, Eyal and Wagner, Allon and others},
  journal={Nature biotechnology},
  volume={40},
  number={9},
  pages={1360--1369},
  year={2022},
  publisher={Nature Publishing Group US New York}
}

@article{vahid2023high,
  title={High-resolution alignment of single-cell and spatial transcriptomes with CytoSPACE},
  author={Vahid, Milad R and Brown, Erin L and Steen, Chlo{\'e} B and Zhang, Wubing and Jeon, Hyun Soo and Kang, Minji and Gentles, Andrew J and Newman, Aaron M},
  journal={Nature biotechnology},
  volume={41},
  number={11},
  pages={1543--1548},
  year={2023},
  publisher={Nature Publishing Group US New York}
}

@article{schmauch2020deep,
  title={A deep learning model to predict RNA-Seq expression of tumours from whole slide images},
  author={Schmauch, Beno{\^\i}t and Romagnoni, Alberto and Pronier, Elodie and Saillard, Charlie and Maill{\'e}, Pascale and Calderaro, Julien and Kamoun, Aur{\'e}lie and Sefta, Meriem and Toldo, Sylvain and Zaslavskiy, Mikhail and others},
  journal={Nature communications},
  volume={11},
  number={1},
  pages={3877},
  year={2020},
  publisher={Nature Publishing Group UK London}
}

@article{pizurica2024digital,
  title={Digital profiling of gene expression from histology images with linearized attention},
  author={Pizurica, Marija and Zheng, Yuanning and Carrillo-Perez, Francisco and Noor, Humaira and Yao, Wei and Wohlfart, Christian and Vladimirova, Antoaneta and Marchal, Kathleen and Gevaert, Olivier},
  journal={Nature Communications},
  volume={15},
  number={1},
  pages={9886},
  year={2024},
  publisher={Nature Publishing Group UK London}
}

@article{haviv2025covariance,
  title={The covariance environment defines cellular niches for spatial inference},
  author={Haviv, Doron and Rem{\v{s}}{\'\i}k, J{\'a}n and Gatie, Mohamed and Snopkowski, Catherine and Takizawa, Meril and Pereira, Nathan and Bashkin, John and Jovanovich, Stevan and Nawy, Tal and Chaligne, Ronan and others},
  journal={Nature Biotechnology},
  volume={43},
  number={2},
  pages={269--280},
  year={2025},
  publisher={Nature Publishing Group US New York}
}

@article{wang2025prediction,
  title={Prediction of cellular morphology changes under perturbations with a transcriptome-guided diffusion model},
  author={Wang, Xuesong and Fan, Yimin and Guo, Yucheng and Fu, Chenghao and Lee, Kinhei and Dallakyan, Khachatur and Li, Yaxuan and Yin, Qijin and Li, Yu and Song, Le},
  journal={Nature Communications},
  volume={16},
  number={1},
  pages={8210},
  year={2025},
  publisher={Nature Publishing Group UK London}
}

@article{lederer2024statistical,
  title={Statistical inference with a manifold-constrained RNA velocity model uncovers cell cycle speed modulations},
  author={Lederer, Alex R and Leonardi, Maxine and Talamanca, Lorenzo and Bobrovskiy, Daniil M and Herrera, Antonio and Droin, Colas and Khven, Irina and Carvalho, Hugo JF and Valente, Alessandro and Dominguez Mantes, Albert and others},
  journal={Nature methods},
  volume={21},
  number={12},
  pages={2271--2286},
  year={2024},
  publisher={Nature Publishing Group US New York}
}

@article{qiu2022mapping,
  title={Mapping transcriptomic vector fields of single cells},
  author={Qiu, Xiaojie and Zhang, Yan and Martin-Rufino, Jorge D and Weng, Chen and Hosseinzadeh, Shayan and Yang, Dian and Pogson, Angela N and Hein, Marco Y and Min, Kyung Hoi Joseph and Wang, Li and others},
  journal={Cell},
  volume={185},
  number={4},
  pages={690--711},
  year={2022},
  publisher={Elsevier}
}

@article{chandrasekaran2024three,
  title={Three million images and morphological profiles of cells treated with matched chemical and genetic perturbations},
  author={Chandrasekaran, Srinivas Niranj and Cimini, Beth A and Goodale, Amy and Miller, Lisa and Kost-Alimova, Maria and Jamali, Nasim and Doench, John G and Fritchman, Briana and Skepner, Adam and Melanson, Michelle and others},
  journal={Nature Methods},
  volume={21},
  number={6},
  pages={1114--1121},
  year={2024},
  publisher={Nature Publishing Group US New York}
}

@article{szalata2024benchmark,
  title={A benchmark for prediction of transcriptomic responses to chemical perturbations across cell types},
  author={Sza{\l}ata, Artur and Benz, Andrew and Cannoodt, Robrecht and Cortes, Mauricio and Fong, Jason and Kuppasani, Sunil and Lieberman, Richard and Liu, Tianyu and Mas-Rosario, Javier A and Meinl, Rico and others},
  journal={Advances in Neural Information Processing Systems},
  volume={37},
  pages={20566--20616},
  year={2024}
}

@article{roohani2025virtual,
  title={Virtual Cell Challenge: Toward a Turing test for the virtual cell},
  author={Roohani, Yusuf H and Hua, Tony J and Tung, Po-Yuan and Bounds, Lexi R and Yu, Feiqiao B and Dobin, Alexander and Teyssier, Noam and Adduri, Abhinav and Woodrow, Alden and Plosky, Brian S and others},
  journal={Cell},
  volume={188},
  number={13},
  pages={3370--3374},
  year={2025},
  publisher={Elsevier}
}

@article{peidli2024scperturb,
  title={scPerturb: harmonized single-cell perturbation data},
  author={Peidli, Stefan and Green, Tessa D and Shen, Ciyue and Gross, Torsten and Min, Joseph and Garda, Samuele and Yuan, Bo and Schumacher, Linus J and Taylor-King, Jake P and Marks, Debora S and others},
  journal={Nature Methods},
  volume={21},
  number={3},
  pages={531--540},
  year={2024},
  publisher={Nature Publishing Group US New York}
}

@article{wei2025perturbase,
  title={PerturBase: a comprehensive database for single-cell perturbation data analysis and visualization},
  author={Wei, Zhiting and Si, Duanmiao and Duan, Bin and Gao, Yicheng and Yu, Qian and Zhang, Zhenbo and Guo, Ling and Liu, Qi},
  journal={Nucleic Acids Research},
  volume={53},
  number={D1},
  pages={D1099--D1111},
  year={2025},
  publisher={Oxford University Press}
}

@article{lotfollahi2019scgen,
  title={scGen predicts single-cell perturbation responses},
  author={Lotfollahi, Mohammad and Wolf, F Alexander and Theis, Fabian J},
  journal={Nature methods},
  volume={16},
  number={8},
  pages={715--721},
  year={2019},
  publisher={Nature Publishing Group US New York}
}

@article{lotfollahi2023predicting,
  title={Predicting cellular responses to complex perturbations in high-throughput screens},
  author={Lotfollahi, Mohammad and Klimovskaia Susmelj, Anna and De Donno, Carlo and Hetzel, Leon and Ji, Yuge and Ibarra, Ignacio L and Srivatsan, Sanjay R and Naghipourfar, Mohsen and Daza, Riza M and Martin, Beth and others},
  journal={Molecular systems biology},
  volume={19},
  number={6},
  pages={e11517},
  year={2023}
}

@article{dong2023causal,
  title={Causal identification of single-cell experimental perturbation effects with CINEMA-OT},
  author={Dong, Mingze and Wang, Bao and Wei, Jessica and de O. Fonseca, Antonio H and Perry, Curtis J and Frey, Alexander and Ouerghi, Feriel and Foxman, Ellen F and Ishizuka, Jeffrey J and Dhodapkar, Rahul M and others},
  journal={Nature methods},
  volume={20},
  number={11},
  pages={1769--1779},
  year={2023},
  publisher={Nature Publishing Group US New York}
}

@article{adduri2025predicting,
  title={Predicting cellular responses to perturbation across diverse contexts with STATE},
  author={Adduri, Abhinav K and Gautam, Dhruv and Bevilacqua, Beatrice and Imran, Alishba and Shah, Rohan and Naghipourfar, Mohsen and Teyssier, Noam and Ilango, Rajesh and Nagaraj, Sanjay and Dong, Mingze and others},
  journal={bioRxiv},
  pages={2025--06},
  year={2025},
  publisher={Cold Spring Harbor Laboratory}
}

@article{wenkel2025txpert,
  title={TxPert: Leveraging Biochemical Relationships for Out-of-Distribution Transcriptomic Perturbation Prediction},
  author={Wenkel, Frederik and Tu, Wilson and Masschelein, Cassandra and Shirzad, Hamed and Eastwood, Cian and Whitfield, Shawn T and Bendidi, Ihab and Russell, Craig and Hodgson, Liam and Mesbahi, Yassir El and others},
  journal={arXiv preprint arXiv:2505.14919},
  year={2025}
}

@article{palma2025predicting,
  title={Predicting cell morphological responses to perturbations using generative modeling},
  author={Palma, Alessandro and Theis, Fabian J and Lotfollahi, Mohammad},
  journal={Nature Communications},
  volume={16},
  number={1},
  pages={505},
  year={2025},
  publisher={Nature Publishing Group UK London}
}

@article{zhang2025cellflux,
  title={CellFlux: Simulating Cellular Morphology Changes via Flow Matching},
  author={Zhang, Yuhui and Su, Yuchang and Wang, Chenyu and Li, Tianhong and Wefers, Zoe and Nirschl, Jeffrey and Burgess, James and Ding, Daisy and Lozano, Alejandro and Lundberg, Emma and others},
  journal={arXiv preprint arXiv:2502.09775},
  year={2025}
}

@article{song2025decoding,
  title={Decoding heterogeneous single-cell perturbation responses},
  author={Song, Bicna and Liu, Dingyu and Dai, Weiwei and McMyn, Natalie F and Wang, Qingyang and Yang, Dapeng and Krejci, Adam and Vasilyev, Anatoly and Untermoser, Nicole and Loregger, Anke and others},
  journal={Nature cell biology},
  pages={1--12},
  year={2025},
  publisher={Nature Publishing Group UK London}
}

@article{nadig2025transcriptome,
  title={Transcriptome-wide analysis of differential expression in perturbation atlases},
  author={Nadig, Ajay and Replogle, Joseph M and Pogson, Angela N and Murthy, Mukundh and McCarroll, Steven A and Weissman, Jonathan S and Robinson, Elise B and O’Connor, Luke J},
  journal={Nature Genetics},
  pages={1--10},
  year={2025},
  publisher={Nature Publishing Group US New York}
}

@article{herzog2017thiol,
  title={Thiol-linked alkylation of RNA to assess expression dynamics},
  author={Herzog, Veronika A and Reichholf, Brian and Neumann, Tobias and Rescheneder, Philipp and Bhat, Pooja and Burkard, Thomas R and Wlotzka, Wiebke and Von Haeseler, Arndt and Zuber, Johannes and Ameres, Stefan L},
  journal={Nature methods},
  volume={14},
  number={12},
  pages={1198--1204},
  year={2017},
  publisher={Nature Publishing Group UK London}
}

@article{erhard2019scslam,
  title={scSLAM-seq reveals core features of transcription dynamics in single cells},
  author={Erhard, Florian and Baptista, Marisa AP and Krammer, Tobias and Hennig, Thomas and Lange, Marius and Arampatzi, Panagiota and J{\"u}rges, Christopher S and Theis, Fabian J and Saliba, Antoine-Emmanuel and D{\"o}lken, Lars},
  journal={Nature},
  volume={571},
  number={7765},
  pages={419--423},
  year={2019},
  publisher={Nature Publishing Group UK London}
}

@article{qiu2020massively,
  title={Massively parallel and time-resolved RNA sequencing in single cells with scNT-seq},
  author={Qiu, Qi and Hu, Peng and Qiu, Xiaojie and Govek, Kiya W and C{\'a}mara, Pablo G and Wu, Hao},
  journal={Nature methods},
  volume={17},
  number={10},
  pages={991--1001},
  year={2020},
  publisher={Nature Publishing Group US New York}
}

@article{mckenna2016whole,
  title={Whole-organism lineage tracing by combinatorial and cumulative genome editing},
  author={McKenna, Aaron and Findlay, Gregory M and Gagnon, James A and Horwitz, Marshall S and Schier, Alexander F and Shendure, Jay},
  journal={Science},
  volume={353},
  number={6298},
  pages={aaf7907},
  year={2016},
  publisher={American Association for the Advancement of Science}
}

@article{chen2024towards,
  title={Towards a general-purpose foundation model for computational pathology},
  author={Chen, Richard J and Ding, Tong and Lu, Ming Y and Williamson, Drew FK and Jaume, Guillaume and Song, Andrew H and Chen, Bowen and Zhang, Andrew and Shao, Daniel and Shaban, Muhammad and others},
  journal={Nature medicine},
  volume={30},
  number={3},
  pages={850--862},
  year={2024},
  publisher={Nature Publishing Group US New York}
}

@article{xu2024whole,
  title={A whole-slide foundation model for digital pathology from real-world data},
  author={Xu, Hanwen and Usuyama, Naoto and Bagga, Jaspreet and Zhang, Sheng and Rao, Rajesh and Naumann, Tristan and Wong, Cliff and Gero, Zelalem and Gonz{\'a}lez, Javier and Gu, Yu and others},
  journal={Nature},
  volume={630},
  number={8015},
  pages={181--188},
  year={2024},
  publisher={Nature Publishing Group UK London}
}

@article{you2024systematic,
  title={Systematic comparison of sequencing-based spatial transcriptomic methods},
  author={You, Yue and Fu, Yuting and Li, Lanxiang and Zhang, Zhongmin and Jia, Shikai and Lu, Shihong and Ren, Wenle and Liu, Yifang and Xu, Yang and Liu, Xiaojing and others},
  journal={Nature methods},
  volume={21},
  number={9},
  pages={1743--1754},
  year={2024},
  publisher={Nature Publishing Group US New York}
}

@article{moshkov2024learning,
  title={Learning representations for image-based profiling of perturbations},
  author={Moshkov, Nikita and Bornholdt, Michael and Benoit, Santiago and Smith, Matthew and McQuin, Claire and Goodman, Allen and Senft, Rebecca A and Han, Yu and Babadi, Mehrtash and Horvath, Peter and others},
  journal={Nature communications},
  volume={15},
  number={1},
  pages={1594},
  year={2024},
  publisher={Nature Publishing Group UK London}
}

@article{fredin2024cell,
  title={Cell Painting-based bioactivity prediction boosts high-throughput screening hit-rates and compound diversity},
  author={Fredin Haslum, Johan and Lardeau, Charles-Hugues and Karlsson, Johan and Turkki, Riku and Leuchowius, Karl-Johan and Smith, Kevin and M{\"u}llers, Erik},
  journal={Nature Communications},
  volume={15},
  number={1},
  pages={3470},
  year={2024},
  publisher={Nature Publishing Group UK London}
}

@article{zhang2025benchmarking,
  title={Benchmarking metabolic RNA labeling techniques for high-throughput single-cell RNA sequencing},
  author={Zhang, Xiaowen and Peng, Mingjian and Zhu, Jianghao and Zhai, Xue and Wei, Chaoguang and Jiao, He and Wu, Zhichao and Huang, Songqian and Liu, Mingli and Li, Wenhao and others},
  journal={Nature Communications},
  volume={16},
  number={1},
  pages={5952},
  year={2025},
  publisher={Nature Publishing Group UK London}
}

@article{yang2024limits,
  title={The limits of fair medical imaging AI in real-world generalization},
  author={Yang, Yuzhe and Zhang, Haoran and Gichoya, Judy W and Katabi, Dina and Ghassemi, Marzyeh},
  journal={Nature Medicine},
  volume={30},
  number={10},
  pages={2838--2848},
  year={2024},
  publisher={Nature Publishing Group US New York}
}

@article{ktena2024generative,
  title={Generative models improve fairness of medical classifiers under distribution shifts},
  author={Ktena, Ira and Wiles, Olivia and Albuquerque, Isabela and Rebuffi, Sylvestre-Alvise and Tanno, Ryutaro and Roy, Abhijit Guha and Azizi, Shekoofeh and Belgrave, Danielle and Kohli, Pushmeet and Cemgil, Taylan and others},
  journal={Nature Medicine},
  volume={30},
  number={4},
  pages={1166--1173},
  year={2024},
  publisher={Nature Publishing Group US New York}
}

@article{kretschmer2025coverage,
  title={Coverage bias in small molecule machine learning},
  author={Kretschmer, Fleming and Seipp, Jan and Ludwig, Marcus and Klau, Gunnar W and B{\"o}cker, Sebastian},
  journal={Nature communications},
  volume={16},
  number={1},
  pages={554},
  year={2025},
  publisher={Nature Publishing Group UK London}
}

@article{joeres2025data,
  title={Data splitting to avoid information leakage with DataSAIL},
  author={Joeres, Roman and Blumenthal, David B and Kalinina, Olga V},
  journal={Nature Communications},
  volume={16},
  number={1},
  pages={3337},
  year={2025},
  publisher={Nature Publishing Group UK London}
}

@article{chevalley2025large,
  title={A large-scale benchmark for network inference from single-cell perturbation data},
  author={Chevalley, Mathieu and Roohani, Yusuf H and Mehrjou, Arash and Leskovec, Jure and Schwab, Patrick},
  journal={Communications Biology},
  volume={8},
  number={1},
  pages={412},
  year={2025},
  publisher={Nature Publishing Group UK London}
}

@article{chen2025giotto,
  title={Giotto Suite: a multiscale and technology-agnostic spatial multiomics analysis ecosystem},
  author={Chen, Jiaji G and Ch{\'a}vez-Fuentes, Joselyn C and O’Brien, Matthew and Xu, Junxiang and Ruiz, Edward C and Wang, Wen and Amin, Iqra and Sheridan, Jeffrey P and Shin, Sujung C and Hasyagar, Sanjana V and others},
  journal={Nature Methods},
  pages={1--13},
  year={2025},
  publisher={Nature Publishing Group US New York}
}

@article{levine2024cell2sentence,
  title={Cell2Sentence: teaching large language models the language of biology},
  author={Levine, Daniel and Rizvi, Syed Asad and L{\'e}vy, Sacha and Pallikkavaliyaveetil, Nazreen and Zhang, David and Chen, Xingyu and Ghadermarzi, Sina and Wu, Ruiming and Zheng, Zihe and Vrkic, Ivan and others},
  journal={BioRxiv},
  pages={2023--09},
  year={2024}
}

@article{gandhi2025tahoe,
  title={Tahoe-x1: Scaling Perturbation-Trained Single-Cell Foundation Models to 3 Billion Parameters},
  author={Gandhi, Shreshth and Javadi, Farnoosh and Svensson, Valentine and Khan, Umair and Jones, Matthew G and Yu, John and Merico, Daniele and Goodarzi, Hani and Alidoust, Nima},
  journal={bioRxiv},
  pages={2025--10},
  year={2025},
  publisher={Cold Spring Harbor Laboratory}
}

@article{miladinovic2025silico,
  title={In silico biological discovery with large perturbation models},
  author={Miladinovic, Djordje and H{\"o}ppe, Tobias and Chevalley, Mathieu and Georgiou, Andreas and Stuart, Lachlan and Mehrjou, Arash and Bantscheff, Marcus and Sch{\"o}lkopf, Bernhard and Schwab, Patrick},
  journal={Nature Computational Science},
  pages={1--12},
  year={2025},
  publisher={Nature Publishing Group US New York}
}

@article{li2025nephrobase,
  title={Nephrobase Cell+: Multimodal Single-Cell Foundation Model for Decoding Kidney Biology},
  author={Li, Chenyu and Ziyadeh, Elias and Sharma, Yash and Dumoulin, Bernhard and Levinsohn, Jonathan and Ha, Eunji and Pan, Siyu and Rao, Vishwanatha and Subramaniyam, Madhav and Szegedy, Mario and others},
  journal={bioRxiv},
  pages={2025--09},
  year={2025},
  publisher={Cold Spring Harbor Laboratory}
}

@article{TejadaLapuerta2025Nicheformer,
  title   = {Nicheformer: a foundation model for single-cell and spatial omics},
  author  = {Tejada-Lapuerta, Alejandro and Schaar, Anna C. and Gutgesell, Robert and others},
  journal = {Nature Methods},
  year    = {2025},
  doi     = {10.1038/s41592-025-02814-z},
  url     = {https://www.nature.com/articles/s41592-025-02814-z},
  note    = {Published online 30 Oct 2025}
}

\end{document}